\documentclass{article}
\usepackage{spconf,amsmath,graphicx}
\usepackage{graphicx}
\usepackage{multirow}
\usepackage[table,xcdraw]{xcolor}
\usepackage{amssymb}
\usepackage{comment}
\usepackage{siunitx}
\usepackage{enumitem}
\setlist{nosep, leftmargin=14pt}

\usepackage{mwe} % to get dummy images

\title{An Efficient Anchor-free Universal Lesion Detection in CT-Scans}
\name{Manu Sheoran$^{\star}$, Meghal Dani$^{\star}$, Monika Sharma, Lovekesh Vig\thanks{$^{\star}$Authors contributed equally}}
\address{TCS Research, New Delhi, India}
\begin{document}
%\ninept
%
\maketitle
\begin{abstract}
%please note !! \textbf{The abstract should appear at the top of the left-hand column of text, about 0.5 inch (12 mm) below the title area and no more than 3.125 inches (80 mm) in length.  Leave a 0.5 inch (12 mm) space between the end of the abstract and the beginning of the main text.  The abstract should contain about 100 to 150}
Existing universal lesion detection (ULD) methods utilize compute-intensive anchor-based architectures which rely on predefined anchor boxes, resulting in unsatisfactory detection performance, especially in small and mid-sized lesions. Further, these default fixed anchor-sizes and ratios do not generalize well to different datasets. Therefore, we propose a robust one-stage anchor-free lesion detection network that can perform well across varying lesions sizes by exploiting the fact that the box predictions can be sorted for relevance based on their center rather than their overlap with the object. Furthermore, we demonstrate that the ULD can be improved by explicitly providing it the domain-specific information in the form of multi-intensity images generated using multiple HU windows, followed by self-attention based feature-fusion and backbone initialization using weights learned via self-supervision over CT-scans. We obtain comparable results to the state-of-the-art methods, achieving an overall sensitivity of $86.05\%$ on the DeepLesion dataset, which comprises of approximately $32K$ CT-scans with lesions annotated across various body organs.

\end{abstract}
\section{Introduction}
\label{sec:intro}
Computer-aided detection/diagnosis (CADe/x) using computed tomography (CT) images has evolved as an emerging field of research, thanks to the tremendous advancements of deep learning techniques in the area of computer vision~\cite{litjens2017survey, kidney, pulm_nodule}.
%The complexity of neural network allows researchers to handle IID (Independent and Identically Distributed) and non-IID datasets. 
Cancer has been one of the most researched and prevalent diseases and the identification of lesions from CT-scans is an important step towards diagnosis. Due to the heterogeneous nature of lesions, their manual analysis and detection is a tedious and error-prone task that requires significant expert knowledge. In the past decade, many efforts have been made towards automated lesion detection but solutions were largely organ-specific focusing on detecting lesions in one of the organs such as liver, kidney, and lungs~\cite{kidney, pulm_nodule}. Recently, the focus has shifted towards developing a Universal Lesion Detector (ULD) which can identify lesions present in different organs from a patient's CT-scan~\cite{yan2019mulan, li2019mvp, 3dce, mla-net, anchorfree-rpn}. DeepLesion~\cite{yan2018deeplesion} is a multi-organ CT-scan dataset, which consists of $32K$ lesions annotated across various organs of the body, available publicly for bench-marking ULD techniques.

\vspace{-1mm}
Prior ULD methods~\cite{3dce, yan2019mulan, li2019mvp, retinanet_improv} have utilized neighboring slice information to provide 3D-context to the network and attention to provide better features for detection by enabling the network to focus on important regions of CT-scans. Yan et al.~\cite{yan2019mulan} proposed MULAN that fuses features from $27$ input slices and jointly trains the network for lesion segmentation and tagging. Some of these works also incorporate novel negative mining techniques to remove false positives and improve the detection. Authors of MELD~\cite{meld} use $4$ different datasets for training and further, use missing annotation matching (MAM) and negative region mining (NRM) for achieving state-of-the-art lesion detection performance on the DeepLesion test-set. We note that the previous ULD methods are anchor-boxe based such as Mask-RCNN~\cite{he2017mask} and FasterRCNN~\cite{ren2015faster}, etc. in which pre-defined fixed anchor-sizes and aspect-ratios are used. This makes it difficult to capture the heterogeneous sizes of lesions present in various body organs of different medical imaging datasets.
Moreover, the anchor-based methods are computationally heavy and quite slow as they require running the detection and classification modules multiple times. This prompted researchers to get rid of anchor-boxes in lesion detection networks~\cite{mla-net, anchorfree-rpn}. Zhang et al.~\cite{anchorfree-rpn} proposed a U-Net based anchor-free ULD network in which each feature map is attached with a detection head. In another work~\cite{mla-net}, authors proposed a compute-heavy multi-layer anchor-free MLANet based on hourglass network and center-to-corner transformation strategy for detecting varied sized lesions. However, both of these anchor-free ULD methods are not the state-of-the-art. 

\begin{figure*}[t]
  \centering
  \includegraphics[width=0.9\linewidth]{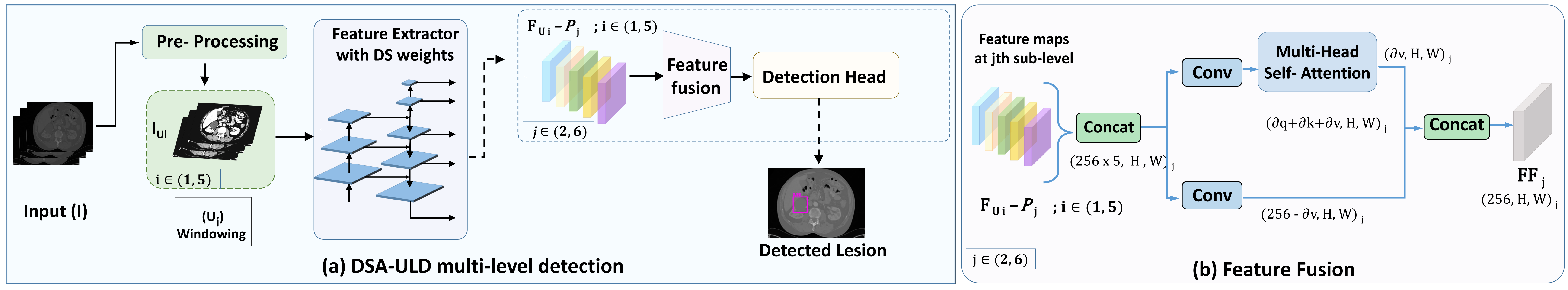}
  \caption{\small{Overview of \emph{DSA-ULD} architecture. (a) An input $I$, consisting of $3$ CT-slices of a patient, is used for generating $5$ multi-intensity images $({I_{U_{i}}})$ with $5$ different HU windows (${U_{1}}$ to ${U_{5}}$). Next, each image is passed through a shared convolutional feature extractor (having domain-specific (DS) weight initialization) with $5$ FPN sub-levels (${P_{2}}$ to ${P_{6}}$) and we extract $5$ feature maps $({F_{Ui}}-{P_{j}})$ for each FPN level $({P_{j}})$. (b) These feature maps at sub-level $({P_{j}})$ are fused together using proposed self-attention module into a single feature map $({FF_{j}})$. Here ${dv}$, ${dk}$, ${dq}$ represent dimensions for Value, Key, and Query matrix. Finally, detection from each sub-level is merged to get the final detection. 
  }}
  \label{fig:arch-dia}
  \vspace{-3mm}
\end{figure*}
\vspace{-1mm}

In this paper, we propose a robust and efficient one-stage anchor-free ULD network that performs at par with state-of-art ULD methods. The concept of a one-stage detector is based on centerness~\cite{tian2019fcos} of objects and predicts all bounding boxes in one go and hence, is computationally efficient and light for clinical deployment. Next, we demonstrate that the performance of automated diagnostic systems can be further improved by imparting extra domain knowledge to the deep networks explicitly. This domain driven information enables deep networks to mimic the diagnostic pattern of doctors by focusing on features or areas where they pay attention while making the final diagnosis. To this end, we provide multi-intensity images highlighting different organs of the body in CT-images using multiple HU windows, fuse the multiple features using self-attention and initialize the backbone of the detection network using medical imaging related weights learned via self-supervision on the DeepLesion~\cite{yan2018deeplesion} dataset. We call our proposed network \emph{DSA-ULD} (Domain-driven Self-attention based Anchor-free ULD).

To summarize, our contributions are as follows:
\begin{itemize}
    \item We propose a one-stage anchor-free ULD network which can efficiently detect lesions present in CT-scans across different organs.
    \item We demonstrate that the feeding of domain information to the deep networks helps to improve detection performance. We named our network \emph{DSA-ULD}: Domain-driven Self-attention based Anchor-free ULD.
    \item We evaluate DSA-ULD on the DeepLesion test-set and achieve better/comparable results with the state-of-the-art ULD methods.
\end{itemize}

\begin{figure*}[t]
  \centering
  \includegraphics[width=0.8\linewidth]{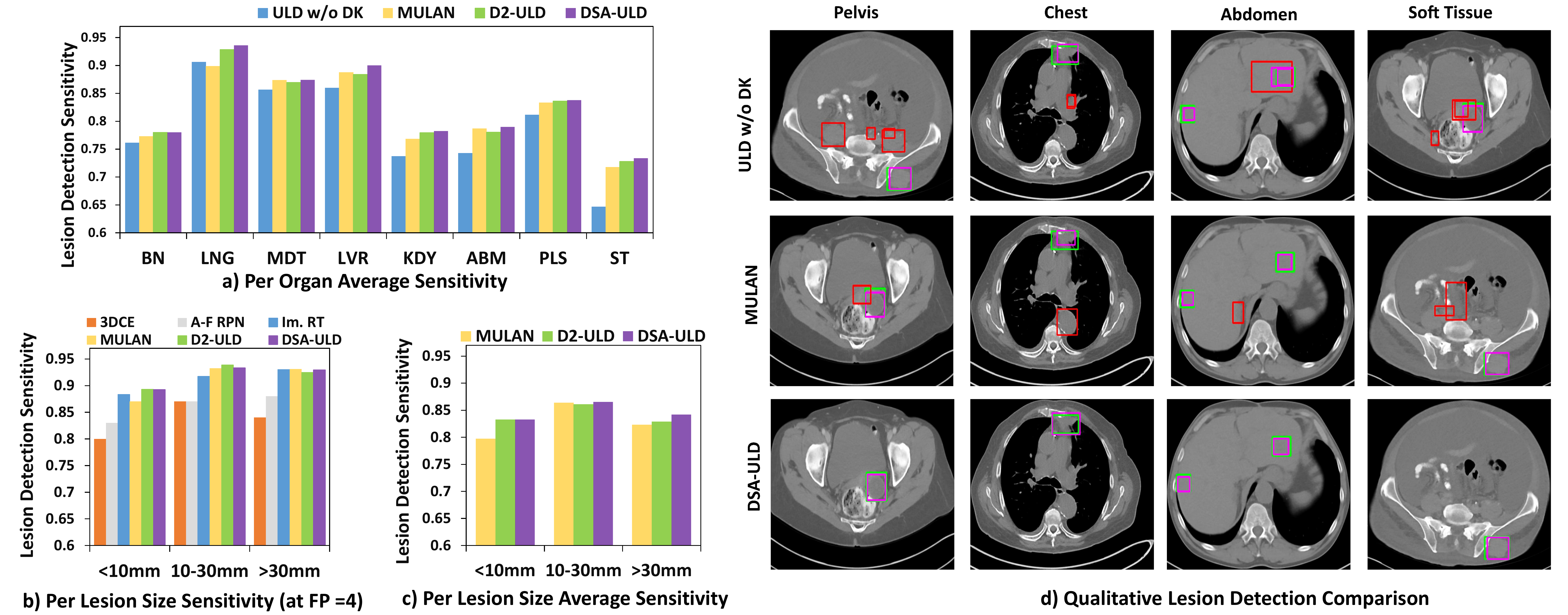}
  \vspace{-2mm}
  \caption{\small{Qualitative and Quantitative Comparison of sensitivity on DeepLesion test-set. Please note that ULD w/o DK represents the anchor-free ULD without domain-knowledge (DK) having $3$ slices as input with only one HU window ($[1024, 4096]$) and without attention-based feature fusion. D2-ULD is a custom-anchor based detectron2 network with DK. Here, BN, LNG, MDT, LVR, KDY, ABM, PLS and ST represent different organs such as bones, lungs, mediastinum, liver, kidney, abdomen, pelvis and soft-tissues, respectively. The green, magenta, and red color boxes represent ground-truth, true-positive (TP), and false-positive (FP) lesion detection, respectively.
  }}
  \label{fig:lesion_size}
  \vspace{-2mm}
\end{figure*}
\vspace{-3mm}

\section{Methodology}
\label{methods}
\vspace{-2mm}
Fig.~\ref{fig:arch-dia} shows our proposed DSA-ULD pipeline. The network takes a 3-channel image $I$ (key slice with one inferior and superior slice) as input and generates $5$ intensity-images $({I_{U_{i}}})$ using $5$ HU-windows $({U_{i}})$ after pre-processing. Next, the features are extracted and fused into $({FF_{j}})$ using a novel feature-fusion strategy based on self-attention which are eventually fed to a detection head. We utilize self-supervised weights and anchor-free protocol to make DSA-ULD generic and robust. Now, we discuss the network in detail:  

\noindent \textbf{Feature Generation:} During manual analysis, a radiologist uses HU windowing to adjust CT intensity values to focus on organs/tissues of interest. Inspired by Masoudi et al.~\cite{masoudi2021quick}, we mimic the radiologist's behaviour in our ULD and
%We mimic the same behavior in our ULD by using multiple HU windows~\cite{xue2012window} to include domain knowledge from CT-scans. Inspired by Masoudi et al.~\cite{masoudi2021quick}, we 
highlight multiple organs of interest with heuristically determined $5$ HU windows~\cite{xue2012window}: $U_1 = {[400,2000]}$, $U_{2,3} = {[-600,1500], [50,350]}$, $U_4 = {[30,150]}$, $U_5 = {[50,400]}$ for bones, chest region including lungs \& mediastinum, abdomen including liver \& kidney, and soft-tissues, respectively. After windowing, 3-channel multi-intensity image $({I_{U_{i}}})$ is passed as input to the ResNeXt-101 shared backbone with feature pyramid network (FPN)~\cite{lin2017featurefpn} based convolutional feature extractor. To further incorporate domain-information in our proposed ULD, we initialize the backbone of our feature extractor with weights learned via self-supervision on the DeepLesion dataset~\cite{grill2020bootstrap}.

\noindent \textbf{Attention Based Feature Fusion:} In order to fuse the feature maps $({F_{ui}-P_{j}})$ containing multi-organ information, one can simply apply a 2D convolution layer which operates only on a local neighborhood. However, recently Vision Transformers~\cite{ViT} have shown remarkable state-of-the-art results across various vision tasks by jointly attending to both spatial and feature sub-spaces with the use of multi-headed self-attention. Therefore, for efficient feature-fusion, we use self-attention which can capture global information across long range dependencies. At each FPN level ${P_{j}}$, $5$ feature maps each having $256$ channels are fed as input to the module. We also use a 2D convolution attention layer ($256-\emph{dv}$ output channels) in parallel with the self-attention module ($\emph{dv}$ output channels) to reduce the computational overhead. Subsequently, the outputs from the two parallel branches are concatenated to obtain the desired number of output channels $(256)$. To reduce computation overhead for attention, we use $2$ heads with the depth of Values matrix as $4$. The dimensions per head for Keys and Values matrix are fixed at $20$. This convolution augmented self-attention allows us to fuse features from different pyramid levels having different resolutions, resulting in robust detection of lesions of different sizes.

\noindent \textbf{Anchor-free One-Stage Detector:} Zhi et al.~\cite{tian2019fcos} proposed a fully convolutional one-stage (FCOS) detector which works on the principle of centerness to reduce the number of low-quality bounding box detections. The overall loss function is:
\vspace{-3mm}

\begin{equation}\label{eq:1}
\begin{split}
     {L(p_{x,y},t_{x,y}) }=  \frac{1}{N_{pos}} \sum_{x,y}^{} L_{cls}(p_{x,y},c^{*}_{x,y})\\ + \frac{\lambda}{N_{pos}}\sum_{x,y}^{} {1_{c^{*}_{x,y}>0}}{ L_{reg}(t_{x,y},t^{*}_{x,y})}
\end{split}
\end{equation}
 In a per-pixel prediction, for each location $(x,y)$ in the feature map ${FF_{j}}$, classification score $p_{x,y}$ is computed followed by regression prediction $t_{x,y}$ for every positive location via an indicator function ${1_{c^*_{x,y}>0}}$. Here, $L_{cls}$ and $L_{reg}$ are the classification focal loss and regression IoU loss for location $(x,y)$, $N_{pos}$ is the no. of positive samples, $\lambda$ is the balance weight, and, ${c^*}$ and ${t^*}$ are ground-truth labels for classification and regression, respectively. Apart from the two conventional detector heads (classification \& regression), there is a third crucial head for centerness. It is based on the idea that low-level regressed boxes that have a skewed feature location in terms of their location inside the box tend to hamper overall detection results. Thus, given the regression targets
$l^*, t^*, r^*$ \&  $b^*$ for a location, the term centerness (as defined below) is trained with binary cross entropy (BCE) loss and added to the loss function defined in Eq.~\ref{eq:1} for the refined results.
\vspace{-2mm}
\begin{equation}
    \small{
    centerness =  \sqrt{ { \frac{min(l^*,r^*)}{max(l^*,r^*)}}  \times {\frac{min(t^*,b^*)}{max(t^*,b^*)}}}  
    }
\end{equation}
% which can directly views locations as training samples (similar to semantic segmentation in FCNs) instead of using anchor boxes as done in anchor-based detectors. By eliminating the anchor boxes, network is made proposal  free which helps to  avoids  the  complicated  computation  related to anchor boxes and also to significantly reduces the number of trainable parameters. Instead of relying on regions proposals, one-stage anchor-free detector can directly leverage as many  as possible foreground sample locations to train the regressor. To further improve the overall performance quality of predicted bounding boxes produced by locations far away from the center of an object a single-layer centerness branch is used in parallel with the classification and bounding box regression layers, which helps suppress the low-quality detected bounding boxes by a large margin.

\section{Experiments and Results}
\label{sec:res}
\vspace{-3mm}
\textbf{Dataset and Metric:}
All the experiments are performed on standard DeepLesion dataset consisting of $32,735$ lesions bounding-box instances on $32,120$ axial key CT slices of $4,427$ unique patients~\cite{yan2018deeplesion}. For a fair comparison, we conduct evaluation on the official test set $(15\%)$, and report sensitivity at various false positives (FPs) per image levels.
\noindent \textbf{Implementation Details:} We utilize weights learned via self-supervised learning (SSL) for FPN backbone with ResNeXt-101 for all the comparison experiments. A $3$-channel image input is generated by taking $3$ slices of a patient's CT-scan (key slice with one superior and inferior neighboring slice). The pre-processing of images include clipping of black borders, windowing~\cite{xue2012window}, and resampling voxel space to $0.8 \times 0.8 \times 2$ mm$^3$. We utilize transformations such as horizontal and vertical flips, resizing and translation to augment the training data.%We augment training data with horizontal and vertical flips, resizing with a ratio of $0.8$ to $1.2$, and translation of $(-8,8)$ pixels in $x$ and $y$ direction. 
The models are trained on NVIDIA Tesla V100 GPUs having $32$GB GPU-memory, with a batch size of $8$. The model is trained until convergence using SGD optimizer with a learning rate and decay-factor of $0.004$ and $10$, respectively.

\noindent \textbf{Comparison with state-of-the-art:} Now, we present comparison results of %\manu{our one stage DSA-ULD method against previously defined one-stage - improved RetinaNet~\cite{retinanet_improv}, Anchor-free RPN~\cite{anchorfree-rpn}, MLANet~\cite{mla-net}, and two-stage - 3DCE~\cite{3dce}, MULAN~\cite{yan2019mulan}, Detectron2 based ULD methods in Table~\ref{tab:comp-sota}} \meghal{not required or can be reformed. just adds text}
ULD methods such as 3DCE~\cite{3dce}, improved RetinaNet~\cite{retinanet_improv}, Anchor-free RPN~\cite{anchorfree-rpn}, MLANet~\cite{mla-net}, MULAN~\cite{yan2019mulan} and Detectron2 based ULD in Table~\ref{tab:comp-sota}. Please note that D2-ULD is the network that is trained using anchor-based Detectron2 backbone in place of anchor-free FCOS and involves domain knowledge such as multi-intensity images, feature-fusion and custom-anchors relevant for lesion-sizes. We do not show a comparison with MELD~\cite{meld} as MELD is trained on $4$ different datasets including DeepLesion and the comparison would not have been fair with our DSA-ULD which is trained on DeepLesion only. We still achieve comparable sensitivity of $86.05\%$ similar to that of MELD ($86.60\%$). This supports our claim that adding extra domain knowledge to the deep networks explicitly alleviates the need for large amounts of heterogeneous training data to learn robust features. As evident in Table~\ref{tab:comp-sota}, we outperform all the prior methods and achieve an average sensitivity of $85.79\%$ with only a few slices of a patient's CT-scan. The average sensitivity is further improved to $86.05\%$ after initializing the backbone of our DSA-ULD with weights learned using self-supervision on the DeepLesion dataset. From Table~\ref{tab:comp-sota}, we also observe that D2-ULD and DSA-ULD give comparable average sensitivity on DeepLesion where D2-ULD uses custom-anchors defined for lesion detection and DSA-ULD is anchor-free. Hence, it can be inferred that anchor-free ULD is more preferable as it does not require very heavy computation while giving equal detection performance.

\begin{table}[!ht]
%\begin{minipage}{\columnwidth}
\begin{center}
\resizebox{\columnwidth}{!}{\begin{tabular}{|l|c|l|c|c|c|c|c|}
\hline
\multirow{2}{*}{\textbf{Method}} & \multirow{2}{*}{\textbf{(S,W)}}  & \multicolumn{4}{c|}{FP (\%)}  & \multirow{2}{*}{\textbf{Avg.}}
\\ \cline{3-6}
& &\textbf{0.5} & \textbf{1.0} & \textbf{2.0} & \textbf{4.0}&
\\\hline
3DCE~\cite{3dce} & (27,1)  & 62.48 & 73.37 & 80.70 & 85.65 & 75.55 \\
%%\hline  
Anchor-Free RPN ~\cite{anchorfree-rpn} & (64,1)  & 68.73 & 77.10 & 83.54 & 88.12 & 79.37 \\
MLANet~\cite{mla-net} & (3,1)  & ---- & 77.10 & 83.0 & 88.30 & ---- \\
%\hline
Improved RetinaNet~\cite{retinanet_improv} & (3,1)& 72.18 & 80.07 & 86.40 & 90.77 & 82.36 \\
MVP Net~\cite{li2019mvp} & (9,3)& 73.83 & 81.82 & 87.60 & 91.30 & 83.64 \\
%\hline
MULAN (w/o tags)~\cite{yan2019mulan} & (27,1)  & 76.10 & 82.50 & 87.50 & 90.90 & 84.33 \\
%\hline
MULAN (w/ tags)~\cite{yan2019mulan} & (27,1)  & 76.12 & 83.69 & 88.76 & 92.30 & 85.22 \\ \hline
%\hline

%MELD(9 slices))~\cite{meld} & 1  & 77.80 & 84.80 & 89.00 & 91.80 & 85.90\\
%\hline
%MELD+MAM+NRM(9 slices)~\cite{meld} & 1  & 78.60 & 85.50 & 89.60 & 92.50 & 86.60 \\ \hline 

D2-ULD (custom anchors) & (3,5) & 75.09 & 83.88 & 89.28   & 92.83  & 85.27   \\ 
D2-ULD + SSL  & (3,5) & 76.07 & 84.31 & \textbf{89.44}  & \textbf{92.94}  & 85.69   \\ \hline
%Ours* (3 Slices) &&&&&&&\\
%(a) D2 & 1 & Default & 67.17 & 77.31 & 84.61 & 90.05 & 79.77 \\
% (b) +attention, x-101 & 3 & Custom & 75.09 & 83.88 & 89.28 & 92.83 & 85.27 \\
(a) \emph{DSA-ULD}* & (3,5) & 77.38 & 84.06& 89.28 & 92.44 & 85.79 \\
(b)\textbf{+SSL} & \textbf{(3,5)} & \textbf{78.30} & \textbf{84.51} & 88.99 & 92.40 & \textbf{86.05}\\

\hline
\end{tabular}}
\vspace{-2mm}
\caption{\small{Sensitivity (\%) Comparison of \emph{DSA-ULD} with previous state-of-the-art methods on DeepLesion~\cite{yan2018deeplesion} test-set. %Sensitivity(\%) at different false-positives (FP) per sub-volume on the volumetric test-set of DeepLesion~\cite{yan2018deeplesion} dataset. 
(S, W) denote no. of slices and HU windows used in each experiment.
}}
\label{tab:comp-sota}
\end{center}
% \bigskip\centering
%\end{minipage}
\vspace{-4mm}
\end{table}
\vspace{-0.5mm}

We also show a comparison of organ-wise average sensitivity, lesion-size wise sensitivity at FP = 4 and lesion-size wise average sensitivity of DSA-ULD with previous methods in Fig.~\ref{fig:lesion_size}(a),(b) and (c), respectively. It is clearly visible that we are able to improve the detection of very small-sized ($< 10mm$) lesions and sensitivity is improved across all the organs of the body. In addition to the above, we present qualitative comparison results (at FP=2) of DSA-ULD in Fig.~\ref{fig:lesion_size}(d) and demonstrate that false positives reduce drastically after the incorporation of domain knowledge in deep networks.

% \begin{table*}[!ht]
%   \begin{tabular}{|l|c|l|c|c|c|c|c|}
% \hline
% \textbf{Method} & \textbf{Windows}  & \textbf{FP@0.5} & \textbf{FP@1.0} & \textbf{FP@2.0} & \textbf{FP@4.0} & \textbf{Average}
% \\\hline
% FCOS\_x101 un-cropped & 1  & 70 & 76 & -- & -- & -- \\
% %%\hline  
% FCOS\_x101 un-cropped & 5  & 75.3 & 82.24 & 87.62 & 91.5 & 84.16 \\
% %\hline
% FCOS\_r50 cropped & 1  & 68.4 & 76.86 & 83 & 87.8 & 79.01 \\
% %\hline
% FCOS\_x101 cropped & 5  & 77.38 & 84.06& 89.28 & 92.44 & 85.79 \\
% %\hline
% FCOS\_x101 cropped + BYOL & 5 & 78.30 & 84.51 & 88.99 & 92.40 & 86.05 \\
% %\hline

% \hline
% \end{tabular}
%   \caption{FCOS, 3 win have DKMA-attention }
% \end{table*}

\vspace{-1mm}
\begin{table}[!ht]
%\begin{minipage}{\columnwidth}
\begin{center}

\setlength{\tabcolsep}{0.6\tabcolsep}% Shrink \tabcolsep by 20%
\begin{tabular}{|c|c|l|c|c|c|}
\hline
\textbf{Sr. No.} &\textbf{HU windows}  &\textbf{Attention}  & \textbf{Avg. Sensitivity} 
\\ \hline
1 &1  &  & 82.29 \\ 
2 &3  &  & 83.71 \\
3 &5  &  & 84.61 \\
%3 & x101 & \checkmark &  & 83.22 \\ 
4 & \textbf{5}  & \textbf{\checkmark}  & \textbf{85.79} \\ 
\hline
\end{tabular}
\vspace{-2mm}
\caption{\small{Ablation studies and average sensitivity (\%) on introducing different no. of HU windows and attention based feature fusion in (\emph{DSA-ULD}) on the DeepLesion test-set.}}
\label{tab:ablation}
\end{center}
\vspace{-4mm}
\end{table}
\vspace{-1mm}

Now, we provide an ablation study on the effect of introducing different numbers of HU windows and attention based feature-fusion in our proposed network DSA-ULD. Table~\ref{tab:ablation} illustrates that when we use our 5 HU windows to generate a multi-intensity input image, we obtain a considerable boost in average sensitivity ($84.61\%$) as compared to using a single HU-window ($82.29\%$). Following this, on applying self-attention based feature fusion, we obtain a further increment in average sensitivity ($85.79\%$). 

% \begin{figure}
%   \centering
%   \includegraphics[width=\linewidth]{pdf/fcos_lesion_bbox.pdf}
%   \caption{\small{Qualitative comparison of \emph{DSA-ULD} and MULAN~\cite{yan2019mulan} (at FP  =$2$) on CT-scans of different body regions. %-(a) Bone, (b) Lungs, (c) Mediastinum, (d) Liver, (e) Kidney, (f)  Abdomen, (g) Pelvis and (h) Soft-tissue . 
%   The green, magenta, and red color boxes represent ground-truth, true-positive (TP), and false-positive (FP) lesion detection, respectively. Please note that ULD w/o DK represents when 3 slices with only one HU window ($[1024, 4096]$) and without attention-based feature fusion are used. We can observe that after incorporating domain knowledge in the form of multi-intensity CT slices, self-attention (i.e., \emph{DSA-ULD}), the number of FP reduced drastically resulting in improved lesion detection performance as compared to MULAN.}}
%   \vspace{-5mm}
%   \label{fig:detection-results}
% \end{figure}

\vspace{-4mm}
\section{Conclusion}
\label{sec:conclusion}
\vspace{-2mm}
We presented an anchor-free one-stage ULD network called DSA-ULD which is also augmented with explicit domain-driven information such as multi-intensity images, feature-fusion using self-attention and self-supervision techniques for efficient and robust lesion detection in CT-scans. We demonstrate that our proposed anchor-free DSA-ULD performs at par with anchor-based lesion detection methods on the DeepLesion test-set while being very simple and computationally-efficient. We also illustrate that the incorporation of domain knowledge in DSA-ULD removes the need of training on heterogeneous datasets. Going forward, we would like to propose a domain-adaptive ULD which can perform effectively on datasets coming from different scanners, domains and hospitals, etc.

%\manu{ We hope that our method can serve as a motivation to design a better and robust ULD which can work across different datasets with multiple organs and handle abnormalities such as motion artifacts, noise etc}

% References should be produced using the bibtex program from suitable
% BiBTeX files (here: strings, refs, manuals). The IEEEbib.bst bibliography
% style file from IEEE produces unsorted bibliography list.
% ------------------------------------------------------------------------- 
\vspace{-2mm}
\small{
\bibliographystyle{IEEEbib}
\bibliography{refs.bib}
}

\end{document}